\documentclass[runningheads]{llncs}
\usepackage[T1]{fontenc}
\usepackage{graphicx}
\usepackage{amsmath}
\usepackage{amssymb}
\usepackage{booktabs}
\usepackage{multirow}
\usepackage{url}
\usepackage{color}
\usepackage[misc]{ifsym}
\usepackage{cite} 
\usepackage{hyperref}

\graphicspath{{./}{./figures/}}

\begin{document}

\title{Decoding Phenotypes: A Framework for Fusing Genomic Language Models and Neuroimaging}
\titlerunning{Decoding Phenotypes}

\author{Tianli Tao\inst{1,2} \and
Ziyang Wang\inst{3} \and
Emma Robinson\inst{1} \and
Rachel Sparks\inst{1} \and
Le Zhang\inst{3}}

\authorrunning{T. Tao et al.}

\institute{School of Biomedical Engineering and Imaging Science, Faculty of Life Science and Medicine, King's College London, UK \\\and
School of Computer Science and Digital Technologies, Aston University, UK\and 
School of Engineering, College of Engineering and Physical Sciences, University of Birmingham, UK
}

\maketitle

\begin{abstract}
Neuroimaging and genetic testing are two important clinical references for nervous system diseases, offering complementary diagnostic information. However, integrating genomic and neuroimaging data for precise disease diagnosis is challenging due to cross-modality heterogeneity. Existing imaging–genetics approaches mainly encode genetic information as hard-coded labels, which lose the local sequence context around disease-associated variants. To address this limitation, we propose GeneFuse, a multimodal learning framework that aligns genetic representations from pre-trained Genomic Language Models (GLMs) with features extracted from images. GeneFuse integrates two components: (1) Genotype-Conditioned Feature Modulation (GCFM), a FiLM-inspired module that uses genomic embeddings to modulate image feature maps; and (2) Uncertainty-aware Genomic Residual Fusion (U-GRF), a fusion strategy that uses imaging-derived predictive uncertainty to gate the contribution of genotypic features. We evaluate GeneFuse on early cognitive decline identification (NC vs. MCI) and dementia screening (NC vs. AD). In the APOE-centered setting, GeneFuse achieves AUROCs of 0.77 and 0.83, outperforming existing imaging–genetics fusion methods. These results indicate that GLM-derived genomic embeddings provide additional information to imaging.

\end{abstract}

\section{Introduction}
Integration of neuroimaging and genetic data is critical for precise diagnosis and understanding of nervous system diseases \cite{parvin2025multimodal}. Neuroimaging captures macroscopic structural phenotypes, such as brain structure and cortical atrophy during aging. Meanwhile, genomics provides complementary information on the underlying disease processes \cite{mashhour2025intersection}. Despite this complementarity, effective data fusion remains challenging due to the substantial heterogeneity in the data representation. Many imaging--genetics fusion approaches represent genetic information as tabular features\cite{li2025imaging}. However, this approach may lose the information about the complex interactions between genomic factors and imaging phenotypes.

Recently, genomic language models (GLMs) have emerged as a promising approach for DNA sequence representation. Foundation models such as the Nucleotide Transformer (NT-v2) \cite{dalla2025nucleotide}, DNABERT \cite{ji2021dnabert}, and HyenaDNA \cite{nguyen2023hyenadna} treat genomic sequences as linguistic structures and leverage self-supervised learning on large-scale genomic datasets to learn context-sensitive representations around disease-associated variants.

Despite the promise of GLM-derived representations, their effective integration with high-dimensional volumetric medical imaging features remains largely unexplored. Existing approaches often depend on handcrafted radiomics or naive late-stage fusion strategies, which may not fully exploit cross-modal interactions between genomics and imaging biomarkers \cite{liu2024leveraging}.  Bridging this gap involves two fundamental challenges. \textit{(1) Semantic misalignment}: Unlike image-text pairs (e.g., CLIP \cite{radford2021learning}), gene-image pairs lack explicit semantic alignment. \textit{(2) Coupling uncertainty}: genotype--phenotype relationships are often weak or indirect, genetic features may not always manifest as observable imaging information due to genetic resilience or environmental factors. Naive fusion strategies carry an inherent risk: when imaging features alone are highly informative, incorporating potentially noisy or loosely coupled genomic signals can degrade predictive performance. 

In this work, we propose a framework that aligns GLM-derived genomic representations with volumetric neuroimaging features. \textbf{Our contributions}: \textbf{(1)}
We adapt feature-wise linear modulation (FiLM) \cite{perez2018film} to align imaging-genomic features through Genotype-Conditioned Feature Modulation (GCFM), which uses genomic embeddings to generate channel-wise affine parameters to modulate volumetric imaging features in an appropriate manner. \textbf{(2)}
We present Uncertainty-aware Genomic Residual Fusion (U-GRF), a lightweight fusion module that uses imaging-derived predictive uncertainty to adaptively gate the contribution of genotypic information. \textbf{(3)} We integrate these components into a network and validate it on Alzheimer's disease (AD) detection tasks using the Alzheimer's Disease Neuroimaging Initiative (ADNI) dataset in single- and multi-locus settings \cite{jack2008alzheimer}. 

\section{Methods}
\subsection{Unimodal Representation and Encoders}

Our framework is designed to process two distinct modalities: volumetric imaging and genomic sequences. In this section, we detail the data construction and encoding networks for each modality.

\textbf{Volumetric Image Encoding.}
To extract high-level anatomical representations from MRI of the brain, we use a 3D TransUNet backbone \cite{chen2021transunet}.  In our implementation, the three encoder stages use channel widths $C_{1/2}=32$, $C_{1/4}=64$, and $C_{1/8}=128$. Each stage contains two $3\times3\times3$ convolutional layers with instance normalization and GELU activation, followed by stride-2 downsampling. A two-layer transformer bottleneck with hidden dimension 256, four attention heads, and an MLP expansion ratio of 4 is applied at the deepest scale. The decoder mirrors the three scales using trilinear upsampling, skip-feature concatenation, and a $3\times3\times3$ convolutional refinement block before each GCFM (see Section 2.2) block. After each GCFM block, globally pooled features from the three scales are concatenated and projected to an imaging latent vector $z_{img} \in \mathbb{R}^{d_{img}}$, where $d_{img}=256$. Given an MRI scan $X_{mri} \in \mathbb{R}^{H \times W \times D}$, the encoder produces multi-scale feature maps $F_s \in \mathbb{R}^{C_s \times H_s \times W_s \times D_s}$ at downsampling scales $s \in \{1/2, 1/4, 1/8\}$, which are used for genotype-conditioned modulation before final aggregation and diagnosis.

\textbf{Semantic Genomic Encoding.}
Unlike traditional association studies that encode Single Nucleotide Polymorphisms (SNPs) as discrete tabular values (i.e., $\{0, 1, 2\}$), we encode local genomic sequence using a pre-trained GLM\cite{dalla2025nucleotide}. This approach preserves nucleotide context around disease-associated variants. 

\textit{Genomic Tokenization \& Context Modeling:} We focus on the APOE locus region as it is a well-established genetic risk factor for late-onset Alzheimer's disease (AD) \cite{liu2013apolipoprotein}. For each subject, we extract a patient-specific DNA sequence window of length $L=1024$ base pairs (bp) centered on the target variant. This window is large enough to encompass the principal proximal regulatory elements. This process yields a patient-specific genomic sentence, which makes the APOE genomic context accessible to the GLM. Although our primary analysis uses APOE, the framework supports multi-locus integration by processing $K$ risk variant sequences independently. In this work, we use permutation-invariant mean pooling ($z_{gene} = \frac{1}{K} \sum_{i=1}^{K} z_{gene}^i$) as the aggregation strategy.

\begin{figure}[!t]
\centering
\includegraphics[width=\textwidth]{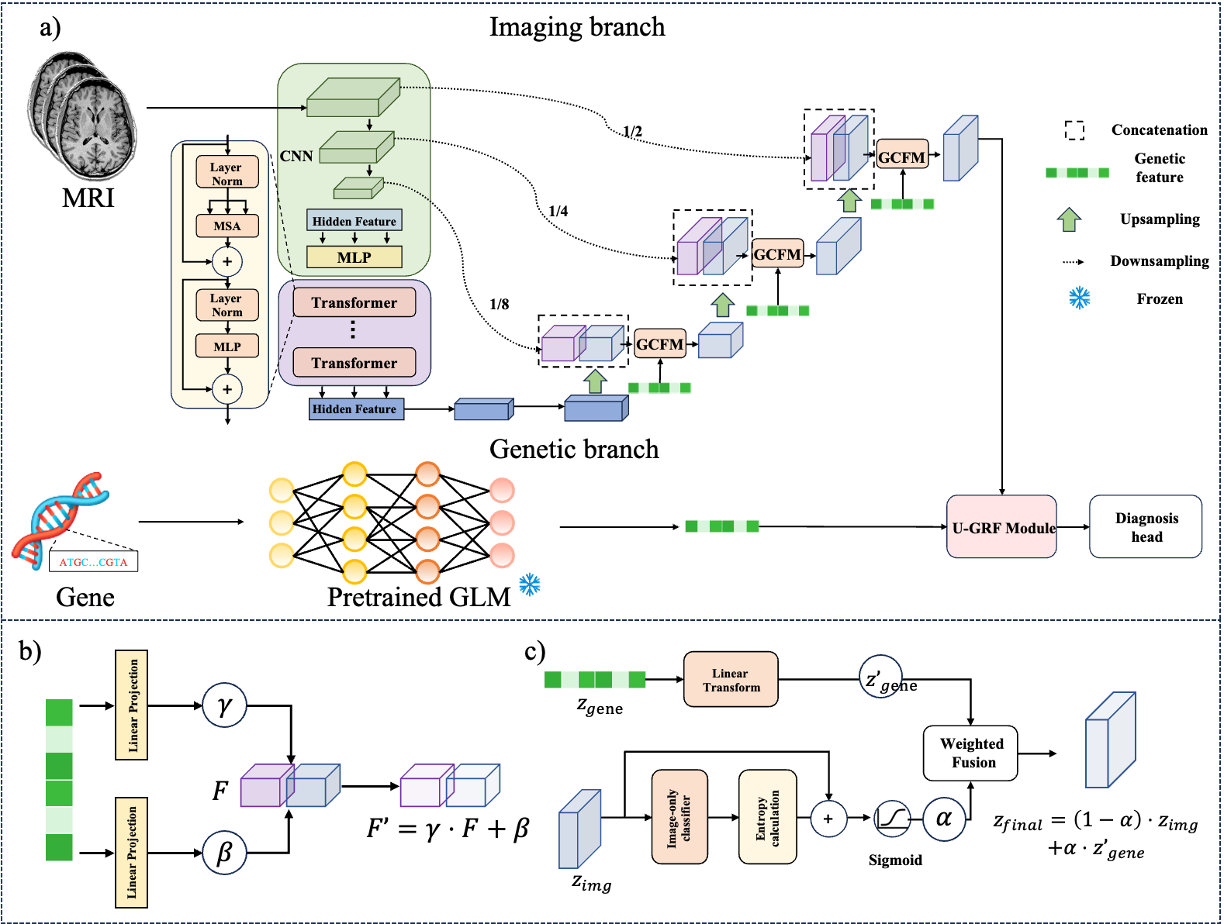} 
\caption{Overview of the proposed GeneFuse framework. (a) The imaging branch extracts multi-scale MRI features with an encoding-decoding backbone, a frozen pre-trained GLM encodes patient-specific genomic encoding. (b) GCFM predicts channel-wise affine parameters ($\gamma$, $\beta$) from genomic embeddings and modulates imaging features by $F' = \gamma \cdot F + \beta$. (c) U-GRF first estimates image-derived uncertainty and then uses the resulting gate $\alpha$ to combine the imaging latent vector $z_{img}$ with the transformed genomic latent vector $z'_{gene}$, producing the fused representation $z_{final}$.}
\label{fig:framework}
\end{figure}

\textit{Genomic Backbone:} To encode each patient-specific sequence window into a dense representation, we employ a pre-trained GLM. In this work, we instantiate the backbone using the Nucleotide Transformer v2 (NT-v2-100M) \cite{dalla2025nucleotide}. We denote the resulting subject-level genomic latent vector as $z_{gene} \in \mathbb{R}^{d_g}$. During training, the parameters of NT-v2 are frozen to preserve the pre-trained genomic representations and reduce computational overhead.

\subsection{Cross-Modal Feature Fusion}
To effectively bridge the gap between imaging and genomic biomarkers, we introduce two fusion modules: Genotype-Conditioned Feature Modulation (GCFM) and Uncertainty-aware Genomic Residual Fusion (U-GRF). These modules work together to ensure genetic information guides visual feature extraction without introducing errors when imaging biomarkers are sufficient for the task.

\textbf{Genotype-Conditioned Feature Modulation (GCFM).} GCFM is an imaging-genetics adaptation of FiLM \cite{perez2018film}. Different from a late FiLM fusion baseline that conditions only the final imaging representation, GCFM injects GLM-derived conditioning into multi-scale spatial feature maps of the 3D volumetric image encoding. This allows genomic information to recalibrate imaging features before the final representation.

In our implementation, GCFM is applied at the $1/2$, $1/4$, and $1/8$ encoder scales after skip-feature concatenation and convolutional refinement. Formally, at a feature stage with channel dimension $C_s$, given the genomic embedding $z_{gene} \in \mathbb{R}^{d_g}$ (where $d_g = 512$ for NT-v2-100M) and the spatial feature map $F_{\text{img}}^s \in \mathbb{R}^{C_s \times H_s \times W_s \times D_s}$, GCFM predicts channel-wise scale ($\gamma_s$) and shift ($\beta_s$) parameters by two separate Multi-Layer Perceptrons (MLPs). Each MLP consists of two linear layers with ReLU activation:
$\text{Linear}(d_g, 256) \to \text{ReLU} \to \text{Linear}(256, C_s)$. Thus:
$\gamma_s = \mathcal{M}_\gamma^s(z_{gene})$,\quad
$\beta_s = \mathcal{M}_\beta^s(z_{gene})$, where $\mathcal{M}_\gamma^s$ and
$\mathcal{M}_\beta^s$ project the genomic latent space to $\mathbb{R}^{C_s}$. The
parameters are applied to the imaging feature map via $F_{\text{img}}^{\prime s} = \gamma_s \odot F_{\text{img}}^s + \beta_s$, where $\odot$ denotes channel-wise multiplication with broadcasting over $H_s, W_s, D_s$. Because the affine parameters are channel-wise and spatially shared, GCFM does not directly predict a voxel-level attention map. Spatially heterogeneous effect arises indirectly from modulating feature channels with different spatially structured activations. The modulated multi-scale imaging features are then aggregated to produce the imaging latent vector $z_{\text{img}}$ for downstream fusion and classification.

\textbf{Uncertainty-aware Genomic Residual Fusion (U-GRF).}
U-GRF is a dynamic fusion strategy that gates the contribution of the genetic features according to the predictive uncertainty of the imaging features to account for the loose coupling between imaging and genomic features.

The fusion process operates in three steps. First, the uncertainty of the image encoder is estimated using the initial logits $l_{img}$ from the imaging branch alone to calculate the predictive entropy $E_{img}$ of the probability distribution $p$, where $p = \text{softmax}(l_{img}), \quad E_{img} = -\sum p \log p$. High entropy indicates diagnostic ambiguity in the visual modality.  Next, a gating coefficient $\alpha$ is computed from $z_{img}$ with $E_{img}$ used to modulate the relative contribution of the genetic encoding $z_{gene}$. This is computed as: $\alpha = \sigma(\text{MLP}_{gate}([z_{img} \oplus E_{img}]))$, where $\sigma$ is the sigmoid function ensuring $\alpha \in [0, 1]$. The genomic branch is therefore modulated by image-derived uncertainty rather than using fused features to decide the extent of fusion. Finally, the joint representation is constructed as: $z'_{gene} = \text{Transform}(z_{gene})$, $ \quad z_{final} = (1-\alpha) \cdot z_{img} + \alpha \cdot z'_{gene}$. When the imaging model is confident (low $E_{img}$), $\alpha$ suppresses $z_{gene}$; when the imaging model is uncertain, $\alpha$ increases the relative contribution of  $z_{gene}$ to the final prediction. The $\text{Transform}(\cdot)$ function is a single linear projection layer, $\text{Linear}(d_g, d_{img})$, which maps $z_{gene}$ from 512 to 256 dimensions.

\section{Experiments}
\textbf{Dataset and Preprocessing.} Data used in this study were obtained from the Alzheimer's Disease Neuroimaging Initiative (ADNI) database \cite{jack2008alzheimer}. We selected subjects with paired T1-weighted (T1w) magnetic resonance imaging (MRI) scans and whole-genome sequencing (WGS). The final dataset contained $N = 182$ subjects spanning three diagnostic groups based on baseline clinical assessments: 52 Normal Controls (NC), 53 subjects with Mild Cognitive Impairment (MCI), and 77 subjects with AD. 

In this work, two clinically motivated diagnosis tasks are considered: NC vs. MCI, corresponding to early identification of cognitive decline, and NC vs. AD, corresponding to dementia screening. All MRI scans underwent a standard pre-processing pipeline, including AC-PC correction, skull stripping, bias field correction, and affine registration to the MNI152 atlas space. Finally, the images were cropped and resized to a resolution of $128 \times 128 \times 128$ and normalized. The genomic pre-processing pipeline followed the pipeline described in \cite{liu2024leveraging}, including variant filtering using \texttt{VCFtools} \cite{danecek2011variant}, genotype phasing using \texttt{Beagle} (v5.5) \cite{browning2021fast} and personal genome construction using \texttt{vcf2diploid} \cite{rozowsky2011alleleseq}.

\textbf{Implementation Details.} Experiments were conducted on the ADNI dataset with stratified 5-fold cross-validation. To prevent data leakage, data splitting was performed at the subject level, so scans and genomic windows from the same subject never appear in both training and test folds. Model training was performed using PyTorch on an NVIDIA 5090 GPU for 50 epochs with AdamW ($lr=10^{-4}, wd=10^{-2}$). The total training objective combines the classification loss with an L1 regularization term on the fusion gate $\alpha$: $\mathcal{L}_{total} = \mathcal{L}_{cls} + \lambda_{reg} \cdot \frac{1}{N} \sum |\alpha|$. Evaluation metrics include AUROC, Accuracy, and F1-Score. Statistical significance for AUROC comparisons is assessed on subject-level out-of-fold predictions using paired DeLong tests for correlated ROC curves, avoiding tests based only on the five fold-level summary values. We compare GeneFuse with unimodal baselines (Image-only, GLM-only), image fusion with scalar SNP encodings, representative imaging--genetics methods (MADDi\cite{golovanevsky2022multimodal}, MMDL\cite{venugopalan2021multimodal}, stage-wise DNN \cite{zhou2019effective}), common fusion strategies (Late Concat, FiLM Fusion \cite{perez2018film}, Cross Attention, Residual), and alternative GLM encoders (DNAGPT \cite{zhang2023dnagpt}, GROVER \cite{sanabria2024dna}, GENA-LM \cite{fishman2025gena}). For the settings of fusion baselines, Late Concat concatenates the final imaging and genomic latent vectors, and Residual adds a fixed-weight transformed genomic residual to the imaging representation. All fusion baselines use the same imaging backbone under the same data split.

\begin{table}
  \centering
  \setlength{\tabcolsep}{1pt}
  \caption{Performance comparison on NC vs. MCI and NC vs. AD tasks. We report mean $\pm$ std. Best results are in bold text. $^\dagger p<0.05$ vs. Image-Only and $^\ddagger p<0.05$ vs. FiLM Fusion using paired DeLong tests on subject-level out-of-fold AUROC predictions.}
  \label{tab:fusion_comparison}
  \setlength{\tabcolsep}{2pt} 
  \resizebox{\textwidth}{!}{%
  \begin{tabular}{l|ccc||ccc}
    \toprule
    & \multicolumn{3}{c||}{\textbf{NC vs. MCI}} & \multicolumn{3}{c}{\textbf{NC vs. AD}} \\
    Method & AUROC & Acc. & F1 & AUROC & Acc. & F1 \\
    \midrule
    \multicolumn{7}{l}{\textit{Baselines}} \\
    GeneFuse (Image-only) & $0.69 \pm 0.09$ & $0.56 \pm 0.07$ & $0.48 \pm 0.10$ & $0.71 \pm 0.05$ & $0.61 \pm 0.07$ & $0.56 \pm 0.13$ \\
    GeneFuse (GLM-only) & $0.63 \pm 0.08$ & $0.50 \pm 0.06$ & $0.45 \pm 0.09$ & $0.71 \pm 0.03$ & $0.58 \pm 0.08$ & $0.54 \pm 0.03$ \\
    Image + SNP scalar& $0.70 \pm 0.09$ & $0.57 \pm 0.07$ & $0.49 \pm 0.11$ & $0.73 \pm 0.09$ & $0.58 \pm 0.05$ & $0.53 \pm 0.06$ \\
    \midrule
    \multicolumn{7}{l}{\textit{Existing Imaging--Genetics Methods}} \\
    Stage-wise DNN \cite{zhou2019effective} & $0.70 \pm 0.02$ & $0.58 \pm 0.05$ & $0.52 \pm 0.08$ & $0.74 \pm 0.07$ & $0.64 \pm 0.05$ & $0.49 \pm 0.06$ \\
    MMDL \cite{venugopalan2021multimodal} & $0.71 \pm 0.08$ & $0.60 \pm 0.07$ & $0.53 \pm 0.07$ & $0.76 \pm 0.10$ & $0.62 \pm 0.08$ & $0.58 \pm 0.12$ \\
    MADDi \cite{golovanevsky2022multimodal} & $0.72 \pm 0.07$ & $0.61 \pm 0.03$ & $0.55 \pm 0.04$ & $0.77 \pm 0.12$ & $0.67 \pm 0.07$ & $0.60 \pm 0.09$ \\
    \midrule
    \multicolumn{7}{l}{\textit{Fusion Strategies}} \\
    Late Concat & $0.67 \pm 0.08$ & $0.52\pm 0.03$ & $0.46 \pm 0.09$ & $0.76 \pm 0.08$ & $0.60 \pm 0.09$ & $0.51 \pm 0.14$ \\
    FiLM Fusion \cite{perez2018film} & $0.72 \pm 0.03$ & $0.59 \pm 0.07$ & $0.52 \pm 0.05$ & $0.73 \pm 0.03$ & $0.61 \pm 0.05$ & $0.54 \pm 0.16$ \\
    Cross-Attention & $0.74 \pm 0.07$ & $0.61 \pm 0.04$ & $0.54 \pm 0.06$ & $0.77 \pm 0.07$ & $0.63 \pm 0.08$ & $0.48 \pm 0.08$ \\
    Residual ($\alpha$=0.05) & $0.73 \pm 0.08$ & $0.60 \pm 0.07$ & $0.53 \pm 0.11$ & $0.76 \pm 0.02$ & $0.62 \pm 0.04$ & $0.56 \pm 0.03$ \\
    \midrule
    \multicolumn{7}{l}{\textit{Ablation Study (Ours w/ NT-V2)}} \\
    GCFM Only & $0.75 \pm 0.06$ & $0.63 \pm 0.06$ & $0.54 \pm 0.09$ & $0.81 \pm 0.06$$^\ddagger$ & $0.65 \pm 0.07$ & $0.52 \pm 0.05$ \\
    U-GRF Only & $0.74 \pm 0.07$ & $0.62 \pm 0.07$ & $0.58 \pm 0.10$ & $0.78 \pm 0.04$ & $0.68 \pm 0.07$ & $0.52 \pm 0.07$ \\
    \textbf{GeneFuse (Full)} & $\mathbf{0.77 \pm 0.05}$ & $\mathbf{0.66 \pm 0.04}$ & $\mathbf{0.60 \pm 0.06}$ & $\mathbf{0.83 \pm 0.05}$$^{\dagger\ddagger}$ & $\mathbf{0.71 \pm 0.03}$ & $\mathbf{0.66 \pm 0.04}$ \\
    \midrule
    \multicolumn{7}{l}{\textit{GLM Comparison (GeneFuse)}} \\
    w/ DNAGPT \cite{zhang2023dnagpt} & $0.75 \pm 0.06$ & $0.63 \pm 0.06$ & $0.57 \pm 0.09$ & $0.79 \pm 0.03$ & $0.64 \pm 0.07$ & $0.62 \pm 0.10$ \\
    w/ GROVER \cite{sanabria2024dna} & $0.74 \pm 0.05$ & $0.62 \pm 0.05$ & $0.56 \pm 0.10$ & $0.80 \pm 0.07$ & $0.65 \pm 0.04$ & $0.55 \pm 0.11$ \\
    w/ GENA-LM \cite{fishman2025gena} & $0.73 \pm 0.08$ & $0.60 \pm 0.02$ & $0.54 \pm 0.11$ & $0.77 \pm 0.06$ & $0.63 \pm 0.08$ & $0.44 \pm 0.16$ \\
    \bottomrule
  \end{tabular}%
  }
\end{table}

\textbf{Experimental Results.} Table \ref{tab:fusion_comparison} reports the APOE-centered comparison. GeneFuse achieves AUROCs of 0.77 (NC vs. MCI) and 0.83 (NC vs. AD), improving over the image-only baseline by $\Delta0.08$ and $\Delta0.12$ AUROC, respectively. These improvements over the image-only backbone are statistically significant under paired DeLong tests on subject-level out-of-fold predictions ($p<0.05$). Image + SNP provides only a small gain, suggesting that scalar SNP encoding is less effective than GLM-derived sequence representations. Compared with existing imaging--genetics baselines, late concatenation, FiLM-style conditioning, cross-attention, and a fixed residual fusion rule, GeneFuse obtains the most favorable overall balance across AUROC, Accuracy, and F1. The ablation study shows complementary roles for the two modules: GCFM improves risk ranking by conditioning multi-scale imaging features on genomic context, while U-GRF improves decision reliability through uncertainty-aware gating.

\begin{figure}[t]
\centering
\includegraphics[width=\textwidth]{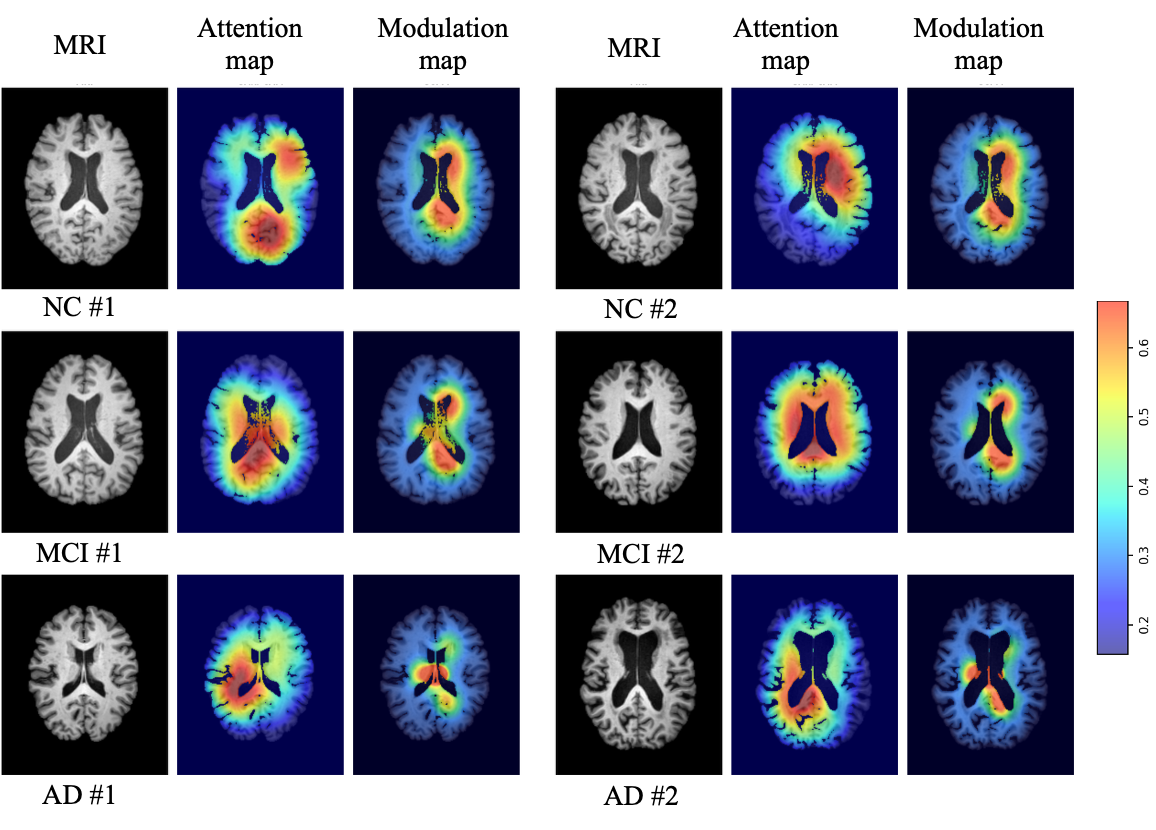} 
\caption{Qualitative comparison between conventional attention map and the proposed GCFM modulation map.}
\label{fig:result2}
\end{figure}

\textbf{Qualitative Visualization.}
To inspect the cross-modal interactions learned by GeneFuse feature modulation effects produced by GCFM are shown in Fig. \ref{fig:result2}. Intermediate feature maps were extracted from the imaging backbone before and after GCFM modulation and compute the modulation intensity as $|F' - F|$ averaged across channels. As illustrated in Fig. 2, genomic conditioning qualitatively produces spatially structured modulation responses around periventricular white matter and medial temporal regions.

\textbf{GCFM Insertion Scale.}
An ablation study was performed on the insertion level of GCFM into the CNN-Transformer U-Net encoder (Table~\ref{tab:gcfm_scale}). Single-scale modulation is competitive with, and in most settings improves over, late FiLM fusion, while using all three encoder scales gives the highest observed performance. This indicates that genomic conditioning benefits from both fine anatomical detail and deeper semantic features.

\begin{table}
  \centering
  \setlength{\tabcolsep}{2pt}
  \caption{Ablation of GCFM insertion scales using NT-v2.}
  \label{tab:gcfm_scale}
  \resizebox{\textwidth}{!}{%
  \begin{tabular}{l|ccc||ccc}
    \toprule
    & \multicolumn{3}{c||}{\textbf{NC vs. MCI}} & \multicolumn{3}{c}{\textbf{NC vs. AD}} \\
    GCFM Scales & AUROC & Acc. & F1 & AUROC & Acc. & F1 \\
    \midrule
    $1/2$ only & $0.72 \pm 0.07$ & $0.56 \pm 0.09$ & $0.49 \pm 0.10$ & $0.78 \pm 0.08$ & $0.64 \pm 0.05$ & $0.59 \pm 0.02$ \\
    $1/4$ only & $0.71 \pm 0.07$ & $0.59 \pm 0.07$ & $0.53 \pm 0.04$ & $0.73 \pm 0.07$ & $0.60 \pm 0.02$ & $0.39 \pm 0.15$ \\
    $1/8$ only & $0.74 \pm 0.03$ & $0.57 \pm 0.08$ & $0.55 \pm 0.08$ & $0.76 \pm 0.08$ & $0.65 \pm 0.08$ & $0.50 \pm 0.05$ \\
    $1/4 + 1/8$ & $0.75 \pm 0.06$ & $0.64 \pm 0.06$ & $0.57 \pm 0.03$ & $0.79 \pm 0.06$ & $0.68 \pm 0.07$ & $0.58 \pm 0.06$ \\
    $1/2 + 1/4 + 1/8$ & $\mathbf{0.77 \pm 0.05}$ & $\mathbf{0.66 \pm 0.04}$ & $\mathbf{0.60 \pm 0.06}$ & $\mathbf{0.83 \pm 0.05}$ & $\mathbf{0.71 \pm 0.03}$ & $\mathbf{0.66 \pm 0.04}$ \\
    \bottomrule
  \end{tabular}%
  }
\end{table}

\textbf{Impact of Genomic Encoders.}
To assess the importance of GLM selection, we compared NT-v2 to alternative genomic encoders (Table~\ref{tab:fusion_comparison}, bottom section). NT-v2 achieves the highest mean performance across both tasks, with modest AUROC gains over other GLMs.

\textbf{Analysis of Different Gene Loci Performance.}
The diagnostic contributions of different AD risk loci are shown in Table \ref{tab:gene_loci}. The single-locus APOE result corresponds to GeneFuse in Table~\ref{tab:fusion_comparison}. Multi-locus integration combining multiple gene loci provides the highest performance in AUROC and accuracy, improving over APOE alone. These results indicate that additional AD-related loci provide complementary information for MRI-genomic fusion.

\begin{table}
  \centering
  \setlength{\tabcolsep}{1pt}
  \caption{Performance comparison across different AD-related gene loci.}
  \label{tab:gene_loci}
  \setlength{\tabcolsep}{2pt} 
  \resizebox{\textwidth}{!}{%
  \begin{tabular}{l|ccc||ccc}
    \toprule
    & \multicolumn{3}{c||}{\textbf{NC vs. MCI}} & \multicolumn{3}{c}{\textbf{NC vs. AD}} \\
    Gene Locus & AUROC & Acc. & F1 & AUROC & Acc. & F1 \\
    \midrule
    Multi-locus & $\mathbf{0.79 \pm 0.04}$ & $\mathbf{0.68 \pm 0.06}$ & $\mathbf{0.62 \pm 0.08}$ & $\mathbf{0.84 \pm 0.02}$ & $\mathbf{0.72 \pm 0.04}$ & $0.63 \pm 0.05$ \\
    APOE & $0.77 \pm 0.05$ & $0.66 \pm 0.04$ & $0.60 \pm 0.06$ & $0.83 \pm 0.05$ & $0.71 \pm 0.03$ & $\mathbf{0.66 \pm 0.04}$ \\
    PVRL2 & $0.72 \pm 0.07$ & $0.60 \pm 0.08$ & $0.53 \pm 0.11$ & $0.77 \pm 0.08$ & $0.64 \pm 0.09$ & $0.48 \pm 0.13$ \\
    ABCA7 & $0.73 \pm 0.05$ & $0.63 \pm 0.05$ & $0.44 \pm 0.13$ & $0.79 \pm 0.08$ & $0.65 \pm 0.08$ & $0.60 \pm 0.12$ \\
    CD33 & $0.72 \pm 0.04$ & $0.61 \pm 0.02$ & $0.53 \pm 0.12$ & $0.78 \pm 0.03$ & $0.64 \pm 0.09$ & $0.49 \pm 0.13$ \\
    APOC1 & $0.68 \pm 0.09$ & $0.59 \pm 0.09$ & $0.42 \pm 0.10$ & $0.71 \pm 0.09$ & $0.62 \pm 0.10$ & $0.56 \pm 0.14$ \\
    \bottomrule
  \end{tabular}%
  }
\end{table}

\section{Conclusion}
GeneFuse shows that GLM-derived risk-locus embeddings can condition MRI features more effectively than scalar SNP encoding or late fusion. The multi-locus results further suggest that AD-related loci beyond APOE contribute addtional diagnostic information. The limitation of this study is that the experiments were conducted using a single cohort, which may limit the generalizability of the findings. In future work, we will extend it to other disease states where genomic–imaging associations may be more diffuse. We will also investigate GLM fine-tuning on larger imaging-genomics datasets to better adapt genomic representations to downstream diagnostic tasks.

\bibliographystyle{splncs04}
\bibliography{references}

\end{document}